\documentclass[10pt,twocolumn,letterpaper]{article}

\usepackage{cvpr}
\usepackage{times}
\usepackage{epsfig}
\usepackage{graphicx}
\usepackage{amsmath}
\usepackage{amssymb}

\usepackage{tabu}
\usepackage{tablefootnote}
\usepackage[pagebackref=true,breaklinks=true,letterpaper=true,colorlinks,bookmarks=false]{hyperref}

\cvprfinalcopy % *** Uncomment this line for the final submission

\ifcvprfinal\fi
\begin{document}

%%%%%%%%% TITLE
\title{Single-Shot Object Detection with Enriched Semantics}

\author{Zhishuai Zhang$^{1}$~~~~~Siyuan Qiao$^{1}$~~~~~Cihang Xie$^{1}$~~~~~Wei Shen$^{1,2}$~~~~~Bo Wang$^{3}$~~~~~Alan L. Yuille$^{1}$\\
Johns Hopkins University$^{1}$~~~Shanghai University$^{2}$~~~Hikvision Research$^{3}$\\
{\tt\small zhshuai.zhang@gmail.com~~~~siyuan.qiao@jhu.edu~~~~cihangxie306@gmail.com}\\
{\tt\small wei.shen@t.shu.edu.cn~~~~wangbo.yunze@gmail.com~~~~alan.yuille@jhu.edu}
}

\maketitle
\thispagestyle{empty}

%%%%%%%%% ABSTRACT
\begin{abstract}
 We propose a novel single shot object detection network named Detection with
 Enriched Semantics (DES).
 Our motivation is to enrich
 the semantics of object detection features within a typical deep detector,
 by a semantic
 segmentation branch and a global activation module.
 The segmentation branch is supervised by weak segmentation
 ground-truth, \ie, no extra annotation is required.
 In conjunction with that,
 we employ a global activation module which
 learns relationship between channels and object classes in a self-supervised
 manner.
 Comprehensive experimental results on both PASCAL VOC and MS COCO detection datasets
 demonstrate the effectiveness of the proposed method.
 In particular, with a VGG16 based DES, we achieve an mAP of 81.7 on VOC2007
 \texttt{test} and an mAP of 32.8 on COCO \texttt{test-dev} with an
 inference speed of 31.5 milliseconds per image on a Titan Xp GPU.
 With a lower resolution version, we achieve an mAP of 79.7 on VOC2007 with
 an inference speed of 13.0 milliseconds per image.
\end{abstract}

%%%%%%%%% BODY TEXT
\section{Introduction}
With the emergence of deep neural networks, computer vision has
been improved significantly in many aspects such as image classification
\cite{he2016deep,huang2017densely,krizhevsky2012imagenet,simonyan2014very,wang2017sort},
object detection~\cite{dai2016r,liu2016ssd,redmon2016you,ren2015faster,tang2017multiple},
and segmentation~\cite{chen2017deeplab, he2017mask, long2015fully}.
Among them, object detection is a fundamental task which has already been
extensively studied.
Currently there are mainly two series of object detection
frameworks: the two-stage frameworks such as Faster-RCNN~\cite{ren2015faster}
and R-FCN~\cite{dai2016r} which extract proposals,
followed by per-proposal classification and regression; and the one-stage
frameworks such as YOLO~\cite{redmon2016you} and SSD~\cite{liu2016ssd}, which apply
object classifiers and regressors in a dense
manner without objectness-based pruning. Both of them do classification
and regression on a set of pre-computed anchors.\par

Previous single shot object detectors, such as SSD, use multiple convolutional
layers to detect objects with different sizes and aspect ratios.
SSD uses a backbone network (\eg, VGG16) to generate a low level detection
feature map. Based on that, several layers of object detection feature maps are built,
learning semantic information in a hierarchal manner. Smaller
objects are detected by lower layers while larger objects are detected by
higher layers.
However, the
low level features usually only capture basic visual patterns without
strong semantic information. This may cause two problems:
small objects may not be detected well,
and the quality of high
level features is also damaged by the imperfect low level features.\par

\begin{figure*}[!tb]
 \includegraphics[width=\linewidth]{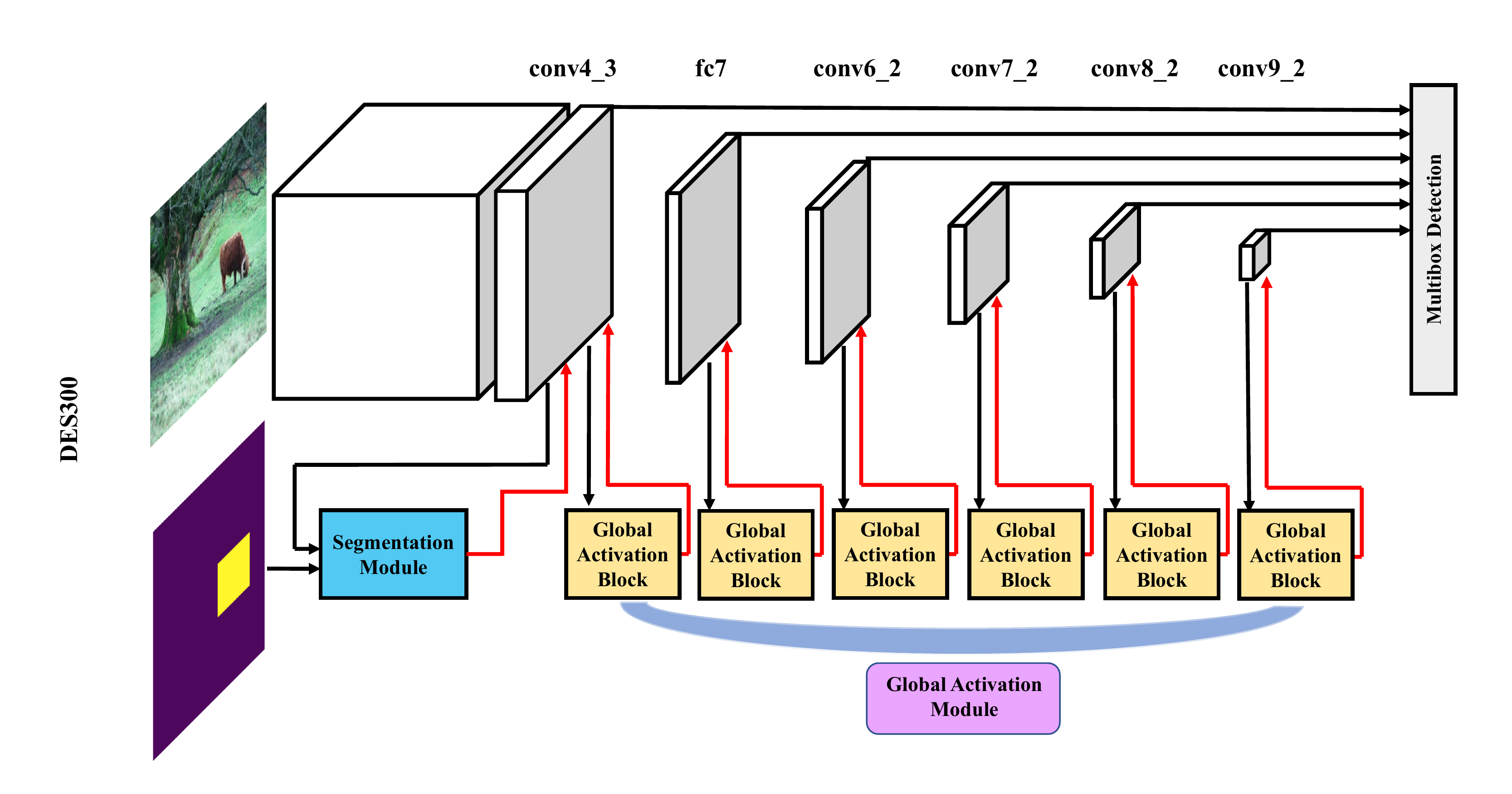}
 \caption{Pipeline for DES:
  the upper half is the
  object detection branch for DES which has six prediction source layers from {\em conv4\_3}
  up to {\em conv9\_2}; the lower half is the
  segmentation branch and the global activation module.
  The segmentation branch is
  added at the first prediction source layer {\em conv4\_3}.
  The global activation module consists of six global activation blocks.
  Those global activation blocks are added at each prediction source layer.
  The black arrows pointed to those modules are
  the input flow, and the red arrows pointed out from those modules are the output flow
  to replace the original feature map.}
 \label{fig:pipeline}
\end{figure*}

In this paper, we aim to address the problem discussed above,
by designing a novel single shot detection network, named
 {\bf Detection with Enriched Semantics} ({\bf DES}),
which consists of two branches, a detection branch and a segmentation
branch. The detection branch is a typical single shot detector,
which takes VGG16 as its backbone,
and detect objects with multiple
object detection feature maps in different layers.
This is shown in the upper part of Figure~\ref{fig:pipeline}.\par

The segmentation branch is used to augment the
low level detection feature map with strong semantic information.
It takes the low level detection feature map as input, to learn semantic segmentation
supervised by bounding-box level segmentation ground-truth. Then it augments the low level
detection features with its semantic meaningful features, as shown in the
left lower part of Figure~\ref{fig:pipeline}.

\begin{figure}[!tb]
 \includegraphics[width=\linewidth]{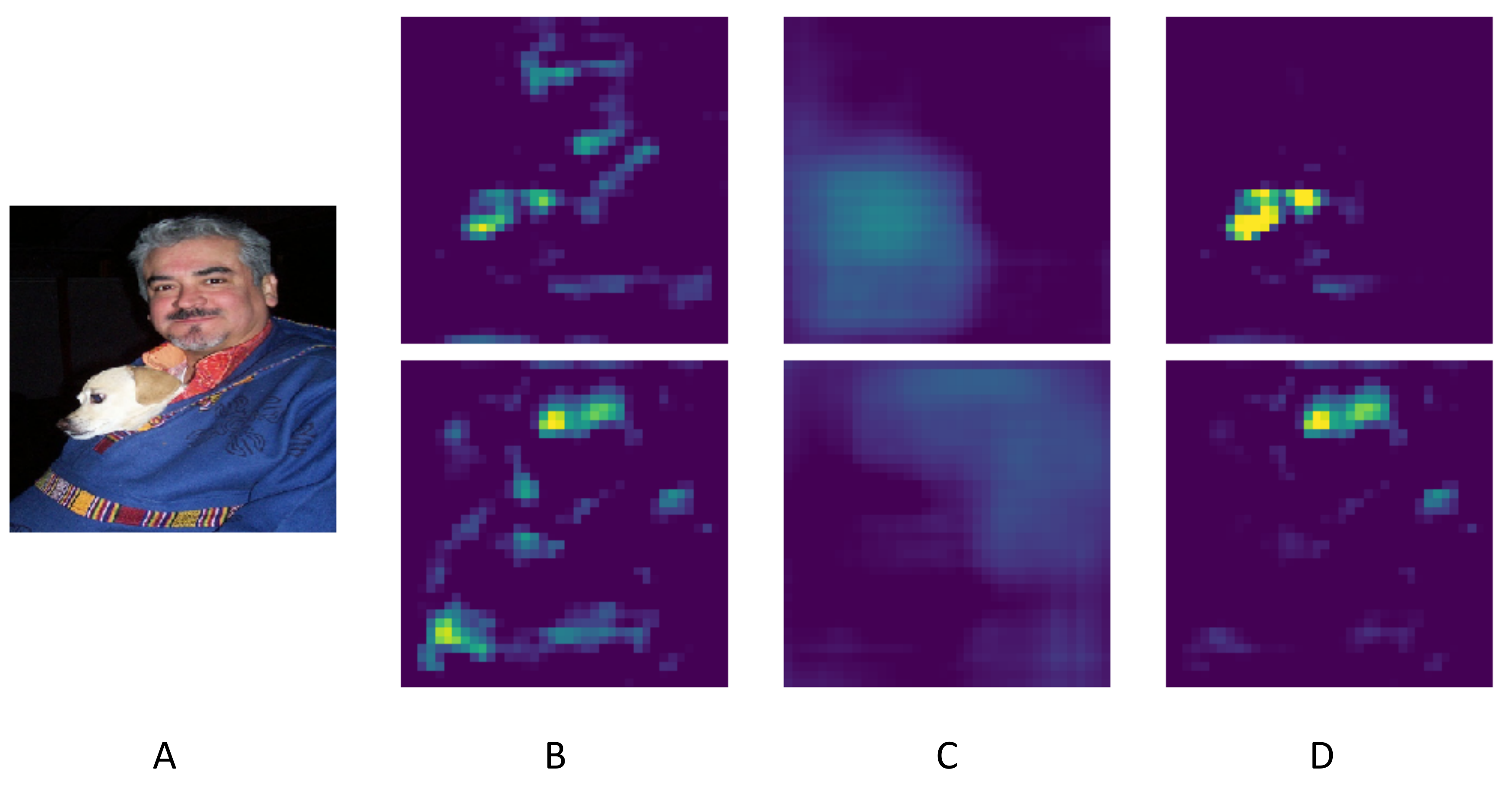}
 \caption{Low level features augmented with semantic meaningful features from
  the segmentation branch. A: original image fed into our detection network.
  B: original low level detection features ($X$) for the input
  image. C: semantic meaningful features ($Z$) from the segmentation branch.
  D: augmented low level features ($X'=X\odot Z$) which is then used in
  the later stages for our detection network. We can see that $X'$ can
  capture both basic visual pattern and high level
  semantic information.}
 \label{fig:activation}
\end{figure}

Figure~\ref{fig:activation} gives an illustration of this semantic augmentation
process. After the original low level features (B) are activated by segmentation
features (C), the augmented low level features (D) can capture
both the basic visual pattern as well as the semantic information of the object.
This can be considered as an attention process, where each channel of the
original low level feature map is activated by a semantically meaningful attention
map, to combine both basic visual pattern and semantically meaningful knowledge.\par

In addition to the segmentation branch attached to the low level detection feature
map,
we also employ a global activation module for higher level detection feature maps.
It consists of several
global activation blocks, as shown in the right lower corner of Figure~\ref{fig:pipeline}.
The global activation block can prune out the location information, and
learn the relationship between channels and object classes in a
self-supervised manner, which increases the semantic information
of the object detection feature maps at higher layers.\par
We summarize our contributions as follows:
\begin{itemize}
 \item We improve the typical deep single shot detectors by enriching
       semantics, with a semantic segmentation
       branch to enhance low level detection features,
       and a global activation module to learn the semantic relationship
       between detection feature channels and object classes for higher
       level object detection features.
 \item We significantly improve the performance compared with popular
       single shot detectors. DES
       achieves an mAP of 81.7 on VOC2007 \texttt{test}
       and mAP of 32.8 on COCO \texttt{test-dev}.
 \item DES is time efficient. With a single Titan Xp GPU,
       it achieves 31.7 FPS,
       and is much faster than competitors like R-FCN and ResNet based
       SSD.
\end{itemize}

\section{Related work}
General object detection is a fundamental task in computer vision and has received
lots of attention.
Almost all the recent object detectors are based on deep networks.
Generally there are two series of object detectors.
The first series is the two-stage detectors.
Some representative examples are
R-CNN~\cite{girshick2014rich}, Fast-RCNN~\cite{girshick2015fast},
Faster-RCNN~\cite{ren2015faster} and R-FCN~\cite{dai2016r}.
These methods first generated
a pool of object candidates, named object proposals, by a separate proposal
generator such as Selective Search~\cite{uijlings2013selective}, Edge
Boxes~\cite{zitnick2014edge} or Region Proposal Network (RPN),
and then did per-proposal classification and bounding box regression.\par

Due to the speed limit of the two-stage frameworks, some research interest has
been attracted by the series of one-stage object detectors,
such as OverFeat~\cite{sermanet2013overfeat},
SSD~\cite{liu2016ssd} and YOLO~\cite{redmon2016you}.
These detectors eliminated the proposal generation, and did object detection
and bounding box regression in a dense manner at different locations and scales.\par

However, all these methods take the object detection as the sole part in the training
phase, without paying close attention to local cues at each position
within the object, which happens to be semantic segmentation.

Semantic segmentation is another important vision task, which requires each pixel
to be assigned to one of classes. General semantic segmentation methods such as DeepLab~\cite{chen2017deeplab}
and fully convolutional network (FCN)~\cite{long2015fully} need per-pixel labelling for the training.
However, it has been shown that weakly annotated training data such as bounding
boxes or image-level labels can also be utilized for semantic segmentation in~\cite{papandreou2015weakly}.\par

We are not the first one to show segmentation information can be leveraged to
help object detection~\cite{gidaris2015object,he2017mask,shrivastava2016contextual}.
Gidaris and Komodakis~\cite{gidaris2015object} used semantic segmentation-aware
CNN features to augment detection features by concatenation at the highest level,
but our work
differs in a way that we put the segmentation information at the lowest detection
feature map, and we use activation instead of concatenation to combine
object detection features and segmentation features.
He \etal \cite{he2017mask}
showed that multi-task training of object detection and instance segmentation
can help to improve the object detection task with extra instance segmentation
annotation, however, we do not consider extra annotation in our work.
Another difference is how the segmentation branch in used.
He \etal \cite{he2017mask} train detection and segmentation in parallel,
but our method uses segmentation features to activate the detection features.\par

Other work such as~\cite{kong2017ron} has been done to improve object detectors
by using top-down architecture to increase the semantic information.
Our work achieves this in a simpler way, which does not involve reverse
connections.

\section{Proposed method}
Detection with Enriched Semantics (DES) is a single-shot object detection network
with three parts: a single shot detection branch, a segmentation branch to
enrich semantics at low level detection layer, and
a global activation module to enrich semantics at higher level detection layers.\par
We use SSD~\cite{liu2016ssd} as our single shot detection branch.
SSD is built on top of a backbone which generates
a low level detection feature map for object detection ({\em conv4\_3} for VGG16). Based
on that, SSD builds a series of feature maps (\ie, {\em conv4\_3} to {\em conv9\_2})
to detect objects of small to
large sizes, in a hierarchical manner, by applying anchors with different
sizes and aspect ratios on these feature maps.
\par
In order to deal with the problems discussed previously, we
employ a segmentation branch to augment low level detection features with
semantic information. This segmentation branch is added at the first prediction
source layer {\em conv4\_3}.
General segmentation algorithms require pixel-level image annotation, but
this is not feasible in the object detection task. Instead, we use bounding-box
level weak segmentation labels to perform supervision for segmentation task.
As shown in the left lower part in Figure~\ref{fig:pipeline},
our segmentation branch takes {\em conv4\_3} as input, represented by the
black arrow pointed from {\em conv4\_3} to segmentation branch.
Then it generates a semantically augmented low level feature map {\em conv4\_3'},
which will be used for detection, represented by the red arrow pointed
from segmentation module to {\em conv4\_3}. By employing segmentation branch,
our network becomes a multi-task learning problem.\par

The feature map generated by the segmentation branch captures high level semantic
information for each local area since the segmentation supervision pushes each
local area to be assigned to one of the classes.\par

At higher level detection layers, the semantic information is already learned
from previous layers; so it is not necessary to employ the segmentation branch
for them. Further, since the resolution is smaller in higher levels,
it will become harder to do the segmentation task based on them. Due to these
reasons, we employ simple global activation blocks, on {\em conv4\_3}
through {\em conv9\_2}, to enrich their semantic
information in a self-supervised manner.

\subsection{Semantic enrichment at low level layer}\label{sec:segmodule}
\begin{figure*}[!tb]
 \includegraphics[width=\linewidth]{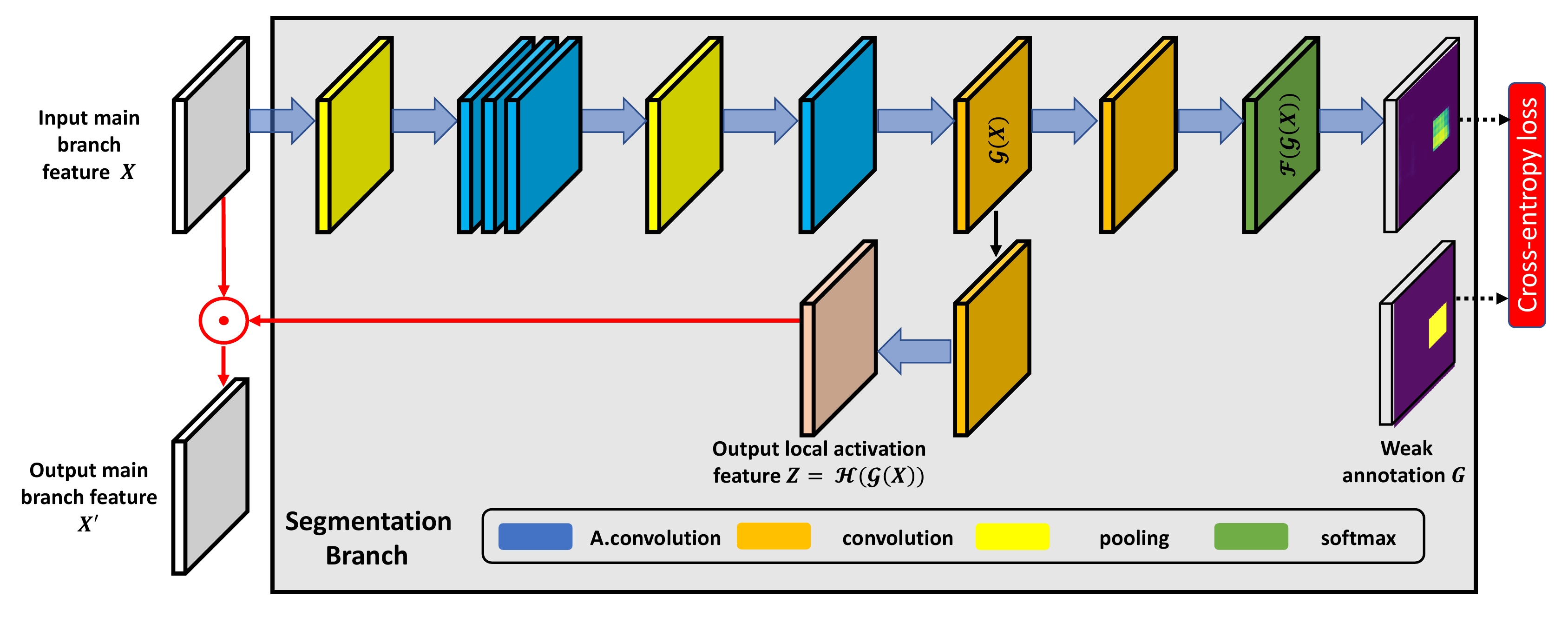}
 \caption{Segmentation branch takes an intermediate feature map from
 object detection branch (\eg~{\em conv4\_3} for SSD300) as input,
 which generates a
 semantically meaningful feature map $Z$ to activate input $X$ to be $X'$. $X'$ is then
 used in the detection branch.}
 \label{fig:segbranch}
\end{figure*}
Semantic enrichment at low level detection
feature layer is achieved by the segmentation branch, which performs
weakly supervised semantic segmentation. It takes the low level detection layer
from the detection branch ({\em conv4\_3} for SSD300) and bounding-box level segmentation
ground-truth as inputs, and generates a semantic meaningful feature map with the same dimension.
Then this feature map is used to activate the input low level detection layer from the detection branch
by element-wise multiplication.\par

Mathematically, let $X\in\mathbb{R}^{C\times H\times W}$ be the low level detection
feature map from the detection branch, $G\in\{0,1,2,\cdots,N\}^{H\times W}$ be the
segmentation ground-truth where $N$ is the number of classes (20 for VOC and 80
for COCO). The segmentation branch computes $Y\in\mathbb{R}^{(N+1)\times H\times W}$
as the prediction of per-pixel segmentation where
$$Y=\mathcal{F}(\mathcal{G}(X))$$ satisfying $$Y\in[0,1]^{(N+1)\times H\times W},
 \sum_{c=0}^{N}Y_{c,h,w}=1\text{.}$$
$\mathcal{G}(X)\in\mathbb{R}^{C'\times H\times W}$ is the intermediate
result which will be further used to generate semantic meaningful feature map:
$$Z=\mathcal{H}(\mathcal{G}(X))\in\mathbb{R}^{C\times H\times W}\text{.}$$
The semantic meaningful feature map $Z$ is then used to activate
the original low level detection feature map $X$ by element-wise multiplication: $X'=X\odot Z$.
where $X'$ is the semantically activated low level detection feature map which conveys
both basic visual patterns and high level semantic information.
$X'$ will replace the original $X$ in the detection branch for object detection.
Figure~\ref{fig:segbranch} gives an illustration of this process.\par

For the segmentation branch, we design a simple network branch mainly composed of
atrous convolutional layers~\cite{chen2017deeplab}.
We add four atrous convolutional layers (noted as `A. convolution' in
Figure~\ref{fig:segbranch}) with $3\times 3$ kernel size after the input feature map $X$.
The first three atrous convolutional layers have a dilation rate of 2 and the last
atrous convolutional layer has a dilation rate of 4.
After that we deploy another $1\times 1$ convolutional layer
to generate $\mathcal{G}(X)$ mentioned above. This intermediate feature map has
two functions: generate segmentation prediction $Y=\mathcal{F}(\mathcal{G}(X))$
and provide high semantic information to activate the input feature map $X'=X\odot
 \mathcal{H}(\mathcal{G}(X))$.
Towards this end, there are two paths attached to $\mathcal{G}(X)$.
The first path ($\mathcal{F}$ path) takes a $1\times 1$ convolution layer with $N+1$ output channels and a
softmax layer to generate the segmentation prediction $Y$.
The second path ($\mathcal{H}$ path) takes another $1\times 1$ convolution layer whose output channel number equals
the channel number of $X$, to generate a semantic meaningful feature map $Z$
in order to activate the feature map in the detection branch by element-wise
multiplication. We show an example of this activation process in Figure~\ref{fig:activation}.
Column A is the input image and column B is one slice of the original low
level object detection feature map $X$.
We can notice that the semantic meaningful feature map $Z$ generated by our segmentation
branch can capture very high level semantic information (the dog or human information).
The final activated feature map $X'$ conveys both
basic visual pattern and high level semantic information.
All these layers keep the size of feature maps unchanged.\par

\begin{figure}[!tb]
 \includegraphics[width=\linewidth]{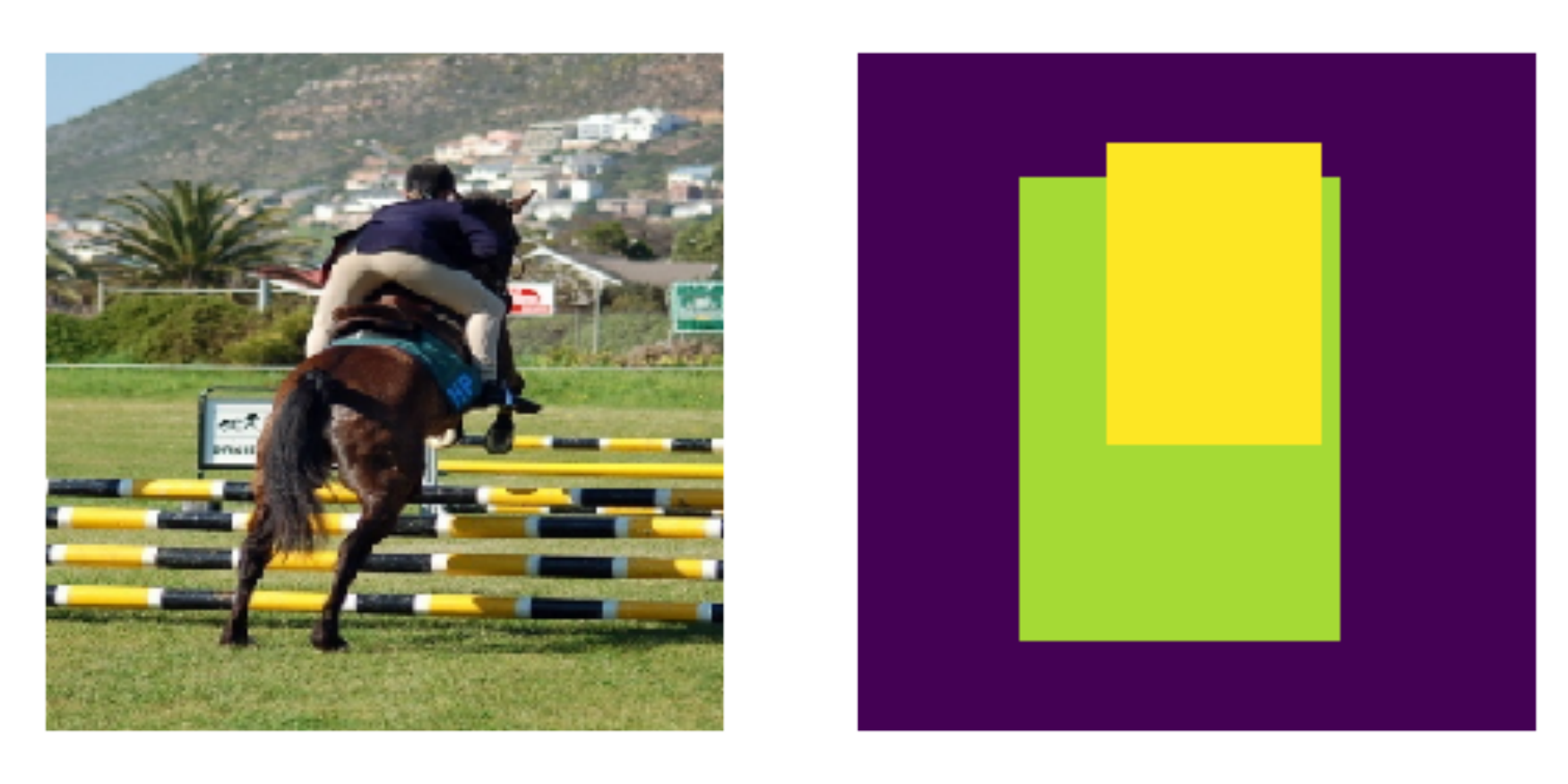}
 \caption{Example of weak segmentation ground-truth. Left: Input image with
  a size of $300\times 300$, with a person and a horse. Right: Segmentation
  ground-truth for the left image, with a size of $38\times 38$;
  the pixels locate in both person and horse bounding-boxes will be
  assigned to person class since its bounding-box is smaller.}
 \label{fig:seggt}
\end{figure}

The final problem is how to generate segmentation ground-truth given only
the object bounding boxes.
The segmentation ground-truth $G$ has the same resolution as the input layer of
segmentation branch ({\em conv4\_3} for SSD300). We use a simple strategy
to generate it: if a pixel $G_{hw}$ locates within
a bounding-box on the image lattice $I$, we assign the label of that bounding-box
to $G_{hw}$; if it locates within more than one bounding-boxes, we choose the label of
the bounding-box with the smallest size; and if it does not locate in any
bounding-box, we assign it to the background class. This strategy guarantees
that there is only one class to be assigned to each pixel in $G$. We show an example
of this weak segmentation ground-truth in Figure~\ref{fig:seggt}.\par

\subsection{Semantic enrichment at higher level layers}
In conjunction with our segmentation branch, we propose
another module named global activation module at higher layers. It contains
several global activation blocks, attached at each object
detection source layer in the detection branch. Global
activation blocks can learn the relationship between
channels and object classes, by eliminating the spatial
information, in a self-supervised manner.\par

The global activation module is shown in the right lower part of
Figure~\ref{fig:pipeline}, which consists of
several global activation blocks attached to each of the object detection source
layer (\ie, {\em conv4\_3} up to {\em conv9\_2} for SSD300).\par
\begin{table*}[!bt]
 \scriptsize
 \centering
 \begin{tabular}{p{2.15cm}p{.75cm}p{.23cm}p{.23cm}p{.23cm}p{.23cm}p{.23cm}p{.23cm}p{.23cm}p{.23cm}p{.23cm}p{.23cm}p{.23cm}p{.23cm}p{.23cm}p{.23cm}p{.23cm}p{.23cm}p{.23cm}p{.23cm}p{.23cm}p{.23cm}p{.23cm}}
  \hline
  % \rowfont{\scriptsize}
  method             & backbone  & mAP       & aero      & bike      & bird      & boat      & bottle    & bus       & car       & cat       & chair     & cow       & table     & dog       & horse     & mbike     & persn     & plant     & sheep     & sofa      & train     & tv        \\ \hline
  Fast~\cite{girshick2015fast}               & VGG16     & 70.0      & 77.0      & 78.1      & 69.3      & 59.4      & 38.3      & 81.6      & 78.6      & 86.7      & 42.8      & 78.8      & 68.9      & 84.7      & 82.0      & 76.6      & 69.9      & 31.8      & 70.1      & 74.8      & 80.4      & 70.4      \\
  Faster~\cite{ren2015faster}             & VGG16     & 73.2      & 76.5      & 79.0      & 70.9      & 65.5      & 52.1      & 83.1      & 84.7      & 86.4      & 52.0      & 81.9      & 65.7      & 84.8      & 84.6      & 77.5      & 76.7      & 38.8      & 73.6      & 73.9      & 83.0      & 72.6      \\
  Faster~\cite{he2016deep}             & ResNet101 & 76.4      & 79.8      & 80.7      & 76.2      & 68.3      & 55.9      & 85.1      & 85.3      & \bf{89.8} & 56.7      & 87.8      & 69.4      & 88.3      & 88.9      & 80.9      & 78.4      & 41.7      & 78.6      & 79.8      & 85.3      & 72.0      \\
  R-FCN~\cite{dai2016r}              & ResNet101 & 80.5      & 79.9      & 87.2      & 81.5      & 72.0      & \bf{69.8} & 86.8      & 88.5      & \bf{89.8} & \bf{67.0} & \bf{88.1} & 74.5      & \bf{89.8} & \bf{90.6} & 79.9      & 81.2      & 53.7      & 81.8      & 81.5      & 85.9      & 79.9      \\
  RON384++~\cite{kong2017ron}           & VGG16     & 77.6      & 86.0      & 82.5      & 76.9      & 69.1      & 59.2      & 86.2      & 85.5      & 87.2      & 59.9      & 81.4      & 73.3      & 85.9      & 86.8      & 82.2      & 79.6      & 52.4      & 78.2      & 76.0      & 86.2      & 78.0      \\ \hline

  Gidaris~\etal~\cite{gidaris2015object}     & VGG16     & 78.2      & 80.3      & 84.1      & 78.5      & 70.8      & 68.5      & 88.0      & 85.9      & 87.8      & 60.3      & 85.2      & 73.7      & 87.2      & 86.5      & 85.0      & 76.4      & 48.5      & 76.3      & 75.5      & 85.0      & 81.0      \\
  Shrivastava~\etal~\cite{shrivastava2016contextual}& VGG16     & 76.4      & 79.3      & 80.5      & 76.8      & 72.0      & 58.2      & 85.1      & 86.5      & 89.3      & 60.6      & 82.2      & 69.2      & 87.0      & 87.2      & 81.6      & 78.2      & 44.6      & 77.9      & 76.7      & 82.4      & 71.9      \\ \hline

  SSD300~\cite{liu2016ssd}             & VGG16     & 77.5      & 79.5      & 83.9      & 76.0      & 69.6      & 50.5      & 87.0      & 85.7      & 88.1      & 60.3      & 81.5      & 77.0      & 86.1      & 87.5      & 84.0      & 79.4      & 52.3      & 77.9      & 79.5      & 87.6      & 76.8      \\
  SSD321~\cite{liu2016ssd}             & ResNet101 & 77.1      & 76.3      & 84.6      & 79.3      & 64.6      & 47.2      & 85.4      & 84.0      & 88.8      & 60.1      & 82.6      & 76.9      & 86.7      & 87.2      & 85.4      & 79.1      & 50.8      & 77.2      & \bf{82.6} & 87.3      & 76.6      \\
  DES300~(Ours)      & VGG16     & 79.7      & 83.5      & 86.0      & 78.1      & 74.8      & 53.4      & 87.9      & 87.3      & 88.6      & 64.0      & 83.8      & 77.2      & 85.9      & 88.6      & \bf{87.5} & 80.8      & 57.3      & 80.2      & 80.4      & \bf{88.5} & 79.5      \\ \hline

  SSD512~\cite{liu2016ssd}             & VGG16     & 79.5      & 84.8      & 85.1      & 81.5      & 73.0      & 57.8      & 87.8      & 88.3      & 87.4      & 63.5      & 85.4      & 73.2      & 86.2      & 86.7      & 83.9      & 82.5      & 55.6      & 81.7      & 79.0      & 86.6      & 80.0      \\
  SSD513~\cite{liu2016ssd}             & ResNet101 & 80.6      & 84.3      & \bf{87.6} & 82.6      & 71.6      & 59.0      & 88.2      & 88.1      & 89.3      & 64.4      & 85.6      & 76.2      & 88.5      & 88.9      & \bf{87.5} & 83.0      & 53.6      & 83.9      & 82.2      & 87.2      & \bf{81.3} \\
  DES512~(Ours)      & VGG16     & \bf{81.7} & \bf{87.7} & 86.7      & \bf{85.2} & \bf{76.3} & 60.6      & \bf{88.7} & \bf{89.0} & 88.0      & \bf{67.0} & 86.9      & \bf{78.0} & 87.2      & 87.9      & 87.4      & \bf{84.4} & \bf{59.2} & \bf{86.1} & 79.2      & 88.1      & 80.5      \\ \hline
\end{tabular}
 \caption{Results on VOC2007 \texttt{test}. The first section contains some representative
  baselines~\cite{dai2016r,girshick2015fast,he2016deep,kong2017ron,ren2015faster}, the second
  section contains other detectors exploiting segmentation information~\cite{gidaris2015object,shrivastava2016contextual},
  the third section contains low resolution SSD and DES, and the last section contains high resolution SSD and DES.
  Note that all these methods are trained on VOC2007 \texttt{trainval} and VOC2012 \texttt{trainval}.}
 \label{tab:voc2007test}
\end{table*}
The global activation block consists of three stages: spatial pooling, channel-wise
learning and broadcasted multiplying.
Formally, given the input
$X\in\mathbb{R}^{C\times H\times W}$, the spatial pooling stage will produce
$Z\in\mathbb{R}^C$ by
$$Z_i=\frac{1}{HW}\sum_{h,w}X_{ihw}$$
and the channel-wise learning stage
will generate the activation feature
$$S=\text{Sigmoid}(W_2\cdot\text{ReLU}
 (W_1Z))\in\mathbb{R}^{C\times 1\times 1}$$
where $W_2\in\mathbb{R}^{C\times C'},W_1\in\mathbb{R}^{C'\times C}$.
In the broadcasted multiplying stage, $S$ is used to
activate $X$ to get $X'\in\mathbb{R}^{C\times H\times W}$ where $X'_{ihw}
 =X_{ihw}\cdot S_i$.
Finally, the $X'$ will replace the original $X$ in the
detection branch. In our experiments, we keep $C'=\frac{1}{4}C$
for all global activation blocks.

This architecture was used for image classification
in~\cite{hu2017squeeze}.
Here we extend it for object detection.

\subsection{Multi-task training}\label{sec:train}
In the training phase, an extra cross-entropy loss function for
segmentation task will
be added in conjunction with the original object detection loss function $L_{\text{det}}(I, B)$ where $I$ is the
image and $B$ is the bounding-box annotation. Our new loss function is formulated
as:
$$L_{\text{seg}}(I, G)=-\frac{1}{HW}\sum_{h,w}\log(Y
 _{G_{h,w},h,w})$$
where $Y\in[0,1]^{(N+1)\times H\times W}$ is the
segmentation prediction, and $G\in\{0,1,2,\cdots,N\}^{H\times W}$ is the
segmentation ground-truth generated by bounding-box annotation, where $N$
is the number of classes excluding background class.\par

By adding the new segmentation loss function to the original detection loss function, the final objective
function we are optimizing is:
$$L(I, B, G)=L_{\text{det}}(I, B)+\alpha L_{\text{seg}}(I, G)$$
where $\alpha$ is a parameter
to balance those two tasks.\par

\section{Experiments}
We present comprehensive experimental results on two main object detection datasets: PascalVOC~\cite{everingham2010pascal}
and MS COCO~\cite{lin2014microsoft}. For PascalVOC, we follow the common split, which uses the
union of VOC2007 \texttt{trainval} and VOC2012 \texttt{trainval} as the training data, and uses
VOC2007 \texttt{test} as the test data. We also show the result on VOC2012
\texttt{test} with the model trained on the union of VOC2007 \texttt{trainvaltest}
and VOC2012 \texttt{trainval}. For COCO, we use a popular split which takes
\texttt{trainval35k}~\cite{bell2016inside} for training, \texttt{minival} for
validation,
and we show results on \texttt{test-dev2017} which is evaluated on the official
evaluation server.\par

For the basic object detection framework, we choose VGG16-based SSD300~\cite{liu2016ssd} and
SSD512 as our single shot detection branch. Note that SSD has been updated
with a new data augmentation trick which boosts the performance with a huge gap.
We follow the latest version of SSD with all those tricks.
The segmentation branch is inserted at the first prediction
source layer, \ie~{\em conv4\_3} for both SSD300 and SSD512. The global activation
module consists of several global activation blocks, 6 for SSD300
and 7 for SSD512, and all of those blocks are added at
each prediction source layer. For the first prediction source layer, the segmentation
branch is inserted before the global activation block.
We follow the SSD training strategy throughout our experiments, and set the
trade-off parameter $\alpha$ to be $0.1$.\par
We will use the terminology `DES300'
and `DES512' to represent our Detection with Enriched Semantics network built
on VGG16-based SSD300 and SSD512 respectively in the rest of our paper.

\subsection{Experiment on VOC}
\begin{table*}[!bt]
 \scriptsize
 \centering
 \begin{tabular}{p{2.15cm}p{.75cm}p{.23cm}p{.23cm}p{.23cm}p{.23cm}p{.23cm}p{.23cm}p{.23cm}p{.23cm}p{.23cm}p{.23cm}p{.23cm}p{.23cm}p{.23cm}p{.23cm}p{.23cm}p{.23cm}p{.23cm}p{.23cm}p{.23cm}p{.23cm}p{.23cm}}
  \hline
  % \rowfont{\scriptsize}
  method             & backbone  & mAP       & aero      & bike      & bird      & boat      & bottle    & bus       & car       & cat       & chair     & cow       & table     & dog       & horse     & mbike     & persn     & plant     & sheep     & sofa      & train     & tv        \\ \hline
  Faster~\cite{he2016deep}             & ResNet101 & 73.8      & 86.5      & 81.6      & 77.2      & 58.0      & 51.0      & 78.6      & 76.6      & 93.2      & 48.6      & 80.4      & 59.0      & 92.1      & 85.3      & 84.8      & 80.7      & 48.1      & 77.3      & 66.5      & 84.7      & 65.6      \\
  R-FCN~\cite{dai2016r}              & ResNet101 & 77.6      & 86.9      & 83.4      & \bf{81.5} & 63.8      & 62.4      & 81.6      & 81.1      & 93.1      & 58.0      & 83.8      & 60.8      & 92.7      & 86.0      & 84.6      & 84.4      & \bf{59.0} & 80.8      & 68.6      & 86.1      & 72.9      \\
  RON384++~\cite{kong2017ron}           & VGG16     & 75.4      & 86.5      & 82.9      & 76.6      & 60.9      & 55.8      & 81.7      & 80.2      & 91.1      & 57.3      & 81.1      & 60.4      & 87.2      & 84.8      & 84.9      & 81.7      & 51.9      & 79.1      & 68.6      & 84.1      & 70.3      \\ \hline

  Gidaris~\etal~\cite{gidaris2015object}     & VGG16     & 73.9      & 85.5      & 82.9      & 76.6      & 57.8      & \bf{62.7} & 79.4      & 77.2      & 86.6      & 55.0      & 79.1      & 62.2      & 87.0      & 83.4      & 84.7      & 78.9      & 45.3      & 73.4      & 65.8      & 80.3      & 74.0      \\
  Shrivastava~\etal~\cite{shrivastava2016contextual} & VGG16     & 72.6      & 84.0      & 81.2      & 75.9      & 60.4      & 51.8      & 81.2      & 77.4      & 90.9      & 50.2      & 77.6      & 58.7      & 88.4      & 83.6      & 82.0      & 80.4      & 41.5      & 75.0      & 64.2      & 82.9      & 65.1      \\ \hline

  SSD300~\cite{liu2016ssd}             & VGG16     & 75.8      & 88.1      & 82.9      & 74.4      & 61.9      & 47.6      & 82.7      & 78.8      & 91.5      & 58.1      & 80.0      & 64.1      & 89.4      & 85.7      & 85.5      & 82.6      & 50.2      & 79.8      & 73.6      & 86.6      & 72.1      \\
  SSD321~\cite{liu2016ssd}             & ResNet101 & 75.4      & 87.9      & 82.9      & 73.7      & 61.5      & 45.3      & 81.4      & 75.6      & 92.6      & 57.4      & 78.3      & 65.0      & 90.8      & 86.8      & 85.8      & 81.5      & 50.3      & 78.1      & 75.3      & 85.2      & 72.5      \\
  DES300~(Ours)\tablefootnote{http://host.robots.ox.ac.uk:8080/anonymous/RCMS6B.html}
                     & VGG16     & 77.1      & 88.5      & 84.4      & 76.0      & 65.0      & 50.1      & 83.1      & 79.7      & 92.1      & 61.3      & 81.4      & 65.8      & 89.6      & 85.9      & 86.2      & 83.2      & 51.2      & 81.4      & \bf{76.0} & 88.4      & 73.3      \\ \hline

  SSD512~\cite{liu2016ssd}             & VGG16     & 78.5      & 90.0      & 85.3      & 77.7      & 64.3      & 58.5      & 85.1      & 84.3      & 92.6      & 61.3      & 83.4      & 65.1      & 89.9      & 88.5      & 88.2      & 85.5      & 54.4      & 82.4      & 70.7      & 87.1      & 75.6      \\
  SSD513~\cite{liu2016ssd}             & ResNet101 & 79.4      & 90.7      & 87.3      & 78.3      & 66.3      & 56.5      & 84.1      & 83.7      & \bf{94.2} & 62.9      & \bf{84.5} & 66.3      & \bf{92.9} & 88.6      & 87.9      & 85.7      & 55.1      & 83.6      & 74.3      & 88.2      & 76.8      \\
  DES512~(Ours)\tablefootnote{http://host.robots.ox.ac.uk:8080/anonymous/OBE3UF.html}
                     & VGG16     & \bf{80.3} & \bf{91.1} & \bf{87.7} & 81.3      & \bf{66.5} & 58.9      & \bf{84.8} & \bf{85.8} & 92.3      & \bf{64.7} & 84.3      & \bf{67.8} & 91.6      & \bf{89.6} & \bf{88.7} & \bf{86.4} & 57.7      & \bf{85.5} & 74.4      & \bf{89.2} & \bf{77.6} \\ \hline
  \end{tabular}
 \caption{Results on VOC2012 \texttt{test}. Note that all methods in this table
  are trained on VOC2007 \texttt{trainvaltest} and VOC2012 \texttt{trainval}, except Gidaris~\etal is trained
  on VOC2007 \texttt{trainval} and VOC2012 \texttt{trainval}.}
 \label{tab:voc2012test}
\end{table*}
For the VOC dataset, we do the training on a machine with 2 Titan Xp GPUs.
To focus on the effectiveness of our DES network, we keep the training settings
used in SSD unchanged.
We first train the model with $\text{lr}=10^{-3}$ for 80k iterations, and then
continue the training with $\text{lr}=10^{-4}$ for 20k iterations and $\text{lr}
 =10^{-5}$ for another 20k iterations. The momentum is fixed to be 0.9 and the
weight decay is set to be 0.0005. Those parameters are aligned with the original
SSD experiments.
We use pre-trained SSD model for VOC to initialize our model,
and initialize the parameters in the first five layers of
segmentation branch with the parameters of
 {\em conv5\_1}, {\em conv5\_2}, {\em conv5\_3}, {\em fc\_6} and {\em fc\_7} in
the detection branch. The rest two convolutional layers of the segmentation
branch are initialized by Xavier initialization~\cite{glorot2010understanding}.
We also do another experiment by resetting all the parameters after {\em conv6\_1} layer
in the detection branch with Xavier initialization. This will lead to similar
results compared with the current setting.\par
The results on VOC2007 \texttt{test} are shown in Table~\ref{tab:voc2007test}.
DES outperforms original SSD on both resolution settings,
and it improves the mAP from 77.5 to 79.7 and
from 79.5 to 81.7 for low and high resolution respectively.
Our VGG16-based model
can even significantly outperform ResNet101-based SSD models, which are much
deeper than VGG16, and this highlights the effectiveness of our method.
\par
Compared with other baselines such as popular two-stage methods and other detector
combined with segmentation, our DES still shows a significant performance
improvement.
For VOC2012 \texttt{test} results shown in Table~\ref{tab:voc2012test}, the same tendency remains.
DES outperforms all the competitors with a large gap.\par
Table~\ref{tab:voc_on_coco} summarizes the results when SSD and DES are fine-tuned from
models trained on COCO. DES outperforms SSD on all test settings with a large
margin. It shows our method can also get benefit from extra training data like the
COCO dataset.
\begin{table}[]
 \centering
\begin{tabular}{l|l|c|c}
\hline
method & backbone & 07~\texttt{test} & 12~\texttt{test} \\ \hline
SSD300~\cite{liu2016ssd} & VGG16    & 79.8                   & 78.5                   \\ \hline
DES300 & VGG16    & 82.7                   & 81.0                   \\ \hline
SSD512~\cite{liu2016ssd} & VGG16    & 83.2                   & 82.2                   \\ \hline
DES512 & VGG16    & 84.3                   & 83.7                   \\ \hline
\end{tabular} \caption{Results on VOC2007~\texttt{test} and VOC2012~\texttt{test} when
 detectors are fine-tuned from models pre-trained on COCO.}\label{tab:voc_on_coco}
\end{table}
\subsection{Experiment on COCO}
We use the similar strategy for COCO task. The DES is implemented from the original
SSD networks which have slightly different default box settings to fit COCO dataset.
The training is conducted on the \texttt{trainvel35k} generated from
COCO \texttt{trainval2014}
dataset. We first train the network with $\text{lr}=10^{-3}$ for 280k iterations,
followed by training with $\text{lr}=10^{-4}$ for 80k iteration and training with
$\text{lr}=10^{-5}$ for another 40k iteration. The momentum is set to be 0.9 and
the weight decay is set to be 0.0005, which are consistent with the original SSD
settings.\par
\begin{table*}[!bt]
 \scriptsize
 \centering
 \begin{tabular}{p{2.12cm}p{1.1cm}p{1.3cm}|p{.53cm}p{.53cm}p{.53cm}|p{.53cm}p{.53cm}p{.53cm}|p{.53cm}p{.53cm}p{.53cm}|p{.53cm}p{.53cm}p{.53cm}}
  \hline
  % \rowfont{\scriptsize}
  method             & backbone  & data        & mAP        & AP50 & AP75 & APsml & APmdm & APlrg & AR1  & AR10 & AR100 & ARsml & ARmdm & ARlrg \\ \hline
  Faster~\cite{ren2015faster}             & VGG16     & trainval    & 21.9       & 42.7 & -    & -     & -     & -     & -    & -    & -     & -     & -     & -     \\
  Faster+++~\cite{he2016deep}          & ResNet101 & trainval    & 34.9       & 55.7 & -    & -     & -     & -     & -    & -    & -     & -     & -     & -     \\
  R-FCN~\cite{dai2016r}              & ResNet101 & trainval    & 29.9       & 51.9 & -    & 10.8  & 32.8  & 45.0  & -    & -    & -     & -     & -     & -     \\
  RON384++~\cite{kong2017ron}           & VGG16     & trainval    & 27.4       & 49.5 & 27.1 & -     & -     & -     & -    & -    & -     & -     & -     & -     \\ \hline

  Shrivastava~\etal~\cite{shrivastava2016contextual} & VGG16     & trainval35k & 27.5       & 49.2 & 27.8 & 8.9   & 29.5  & 41.5  & 25.5 & 37.4 & 38.3  & 14.6  & 42.5  & 57.4  \\ \hline
  SSD300~\cite{liu2016ssd}             & VGG16     & trainval35k & 25.1       & 43.1 & 25.8 & 6.6   & 25.9  & 41.4  & 23.7 & 35.1 & 37.2  & 11.2  & 40.4  & 58.4  \\
  SSD321~\cite{liu2016ssd}             & ResNet101 & trainval35k & 28.0       & 45.4 & 29.3 & 6.2   & 28.3  & 49.3  & 25.9 & 37.8 & 39.9  & 11.5  & 43.3  & 64.9  \\
  DES300~(Ours)      & VGG16     & trainval35k & 28.3       & 47.3 & 29.4 & 8.5   & 29.9  & 45.2  & 25.6 & 38.3 & 40.7  & 14.1  & 44.7  & 62.0  \\ \hline

  SSD512~\cite{liu2016ssd}             & VGG16     & trainval35k & 28.8       & 48.5 & 30.3 & 10.9  & 31.8  & 43.5  & 26.1 & 39.5 & 42.0  & 16.5  & 46.6  & 60.8  \\
  SSD513~\cite{liu2016ssd}             & ResNet101 & trainval35k & 31.2       & 50.4 & 33.3 & 10.2  & 34.5  & 49.8  & 28.3 & 42.1 & 44.4  & 17.6  & 49.2  & 65.8  \\
  DES512~(Ours)      & VGG16     & trainval35k & 32.8       & 53.2 & 34.6 & 13.9  & 36.0  & 47.6  & 28.4 & 43.5 & 46.2  & 21.6  & 50.7  & 64.6  \\ \hline
\end{tabular}
 \caption{Results on COCO \texttt{test-dev}. `sml', `mdm' and `lrg' stand for small, medium
  and large respectively, and `mAP', `AP50' and `AP75' mean average precision of IOU
  =0.5:0.95, IOU=0.5 and IOU=0.75 respectively. \texttt{trainval35k} is obtained by removing
  the 5k \texttt{minival} set from \texttt{trainval}.}
 \label{tab:coco}
\end{table*}
Similar to our methods used for VOC, we use the pre-trained SSD model for COCO to
initialize our parameters, and use weights in $conv5\_1$, $conv5\_2$, $conv5\_3$,
$fc\_6$ and $fc\_7$ to initialize the first five layers in the segmentation branch.
However, different from VOC, we
find that resetting weights after $conv6\_1$ is crucial for good performance,
and we can only get a small improvement around 0.2 for AP@0.5 if we keep those
weights after $conv6\_1$ same as the SSD pre-trained model.\par
We report results on COCO \texttt{test-dev2017} with 20288 images from the official
evaluation server deployed on CodaLab
in Table~\ref{tab:coco}. Compared with our baseline SSD, our DES can
provide huge improvement on all of the metrics. For the low resolution version
(the third section in the table), we can achieve a relative
improvement of 12.7\% for mAP compared with baseline SSD300, from 25.1 to 28.3,
and a significant relative improvement of 28.8\% for small objects. For the high
resolution version (the fourth section in the table), DES
can improve the baseline from 28.8 to 32.8. Our DES can also outperform
SSD based on ResNet101, which is deeper and much slower.\par
We can find that DES performs very good on small objects, outperforms at least
27.5\% relatively compared with all other competitors which report
performance on small objects. Although DES512 outperforms SSD512 based
on VGG16 for detecting large objects, it is slightly worse than SSD513 based
on ResNet101. We argue that SSD513 can benefit from ResNet101
which is much deeper, to detect large objects.
\subsection{Discussion}
\subsubsection{Architecture ablation and diagnosis}\label{sec:ablation}
To further understand the effectiveness of our two extra modules, we do experiments
with different settings and report the results in Table~\ref{tab:ablation} on the
VOC2007 \texttt{test} dataset based on DES300.\par
As can be seen from Table~\ref{tab:ablation}, the global activation module (G)
can improve the performance by 0.6, which shows the effectiveness of
global activation with global activation features.
With the segmentation branch (S)
added, the performance can be further improved with a large margin,
which confirms our intuition that segmentation can be used to help object
detection, and introducing high level semantic knowledge to the early stage of
the detection network can contribute to a stronger object detector.\par
\begin{table}[]
 \centering
 \begin{tabular}{l|l}
  \hline
  method                    & mAP       \\ \hline
  SSD300                    & 77.5      \\ \hline
  SSD300+G                  & 78.1      \\ \hline
  SSD300+G+S~($\alpha=0.0$) & 79.4      \\ \hline
  SSD300+G+S~($\alpha=0.1$) & \bf{79.7} \\ \hline
  SSD300+G+S~($\alpha=1.0$) & 78.6      \\ \hline
  SSD300+G+S~(in parallel)  & 78.2      \\ \hline
  SSD300+G+DeeperVGG16      & 77.6      \\ \hline
 \end{tabular}
 \caption{Ablation result evaluated on VOC2007 \texttt{test} dataset.
  G stands for the global activation module and S
  stands for the segmentation branch. $\alpha$ is the hyper-parameter
  controlling the tradeoff between segmentation loss and detection loss
  discussed in Section~\ref{sec:train}.}\label{tab:ablation}
\end{table}
Another ablation study conducted is the weight of the segmentation loss.
To do this, we train our DES network for VOC2007 \texttt{test} task with
different $\alpha$'s, \ie, 0, 0.1 and 1. This hyper-parameter plays an important
role for balancing object detection and segmentation tasks. Experiments shows that
$\alpha=0.1$ yields the best performance, 0.3 better than $\alpha=0$
(eliminating segmentation loss) and 1.1 better than $\alpha=1$
(taking the tasks of object detection and segmentation equally important).
This means the supervision over the segmentation task is important in our
segmentation branch. But it should take less weight
since the final task is object detection instead of segmentation,
otherwise the segmentation module would lean toward the segmentation task
too much and hurt the detection performance.\par
To further justify the effectiveness of our segmentation branch architecture, we
conduct another two experiments. In the first experiment, we mimic Mask-RCNN~\cite{he2017mask}
by training segmentation and detection branches in parallel, in stead of using
segmentation features to activate low level detection features. The improvement
is very small (mAP of 78.2 as shown in the 6-th row in Table~\ref{tab:ablation})
and we believe the activation process is very important to improve
detection features, and since our weak segmentation ground-truth does not contain
extra information, it will not improve the performance significantly if trained in parallel.
As a side evidence, we train two versions of Mask-RCNN, with no segmentation supervision
and with weak segmentation supervision respectively, and the performance only
goes up by a small amount of 0.6 on COCO \texttt{minival}. This indicates that
Mask-RCNN cannot get a huge benefit from the weak segmentation supervision
trained in parallel, and confirms our observation on DES.
The second experiment we
do is removing the segmentation loss and the activation
process. Then the $Z=\mathcal{H}(\mathcal{G}(X))$ is directly used by the object
detection branch. This modification keeps the number of parameters introduced by our
segmentation branch, and can be regraded as a `deeper VGG16' with more parameters
as the backbone.
This architecture achieves an mAP of 77.6, which is much lower than our DES.
This means the architecture of our segmentation branch is crucial, and the performance
can be worse by naively adding more layers and parameters.

\subsubsection{Inference Speed}\label{sec:speed}
To quantitatively evaluate the inference speed, we run DES, SSD, as well as R-FCN,
on our machine with an nVIDIA Titan Xp GPU to compare the
speed fairly.\par
\begin{table}[!tb]
 % \small
 \centering
\resizebox{\linewidth}{!}{
 \begin{tabular}{llcccc}
  \hline
  % \rowfont{\scriptsize}
  method & backbond  & mAP  & time~(ms/img) & FPS  & batchsize \\\hline
  R-FCN~\cite{dai2016r}  & ResNet101 & 80.5 & 89.6          & 11.2 & 1         \\\hline
  SSD300~\cite{liu2016ssd} & VGG16     & 77.5 & 9.2           & 109.3 & 8         \\
  SSD321~\cite{liu2016ssd} & ResNet101 & 77.1 & 33.2          & 30.2 & 8         \\
  DES300 & VGG16     & 79.7 & 13.0          & 76.8 & 8         \\\hline
  SSD512~\cite{liu2016ssd} & VGG16     & 79.5 & 18.6          & 53.8 & 8         \\
  SSD513~\cite{liu2016ssd} & ResNet101 & 80.6 & 61.6          & 16.2 & 8         \\
  DES512 & VGG16     & 81.7 & 31.5          & 31.7 & 8         \\\hline
\end{tabular}}
 \caption{Inference Speed of two-shot baseline R-FCN and single-shot SSD
  and DES under different resolutions. Here we report the mAP on VOC2007 \texttt{test} dataset
  in the mAP column, the time spent for inferring one image in milliseconds in the
  time column, as well as the number images processed within one second in the
  FPS column.}
 \label{tab:speed}
\end{table}

All results are shown in Table~\ref{tab:speed}. Note that to make comparison fair,
we keep the batchsize to be the same in each comparison group (\ie low resolution
group based on SSD300 and the high resolution group based on SSD512).
For ResNet101 based SSD321 and SSD513, we remove the batch normalization layer at the test
time to reduce the run time and memory consumption following~\cite{fu2017dssd}.\par

Our method is slower than original VGG16-based SSD due to our extra modules,
however, DES is faster than ResNet101-based SSD with a large margin, and
outperforms it in the meantime. DES300 has an FPS of 76.8 with mAP of 79.7,
while DES512 achieves higher mAP with more inference time.

\subsubsection{Detection examples}\label{sec:demo}
\begin{figure}[!tb]
 \includegraphics[width=\linewidth]{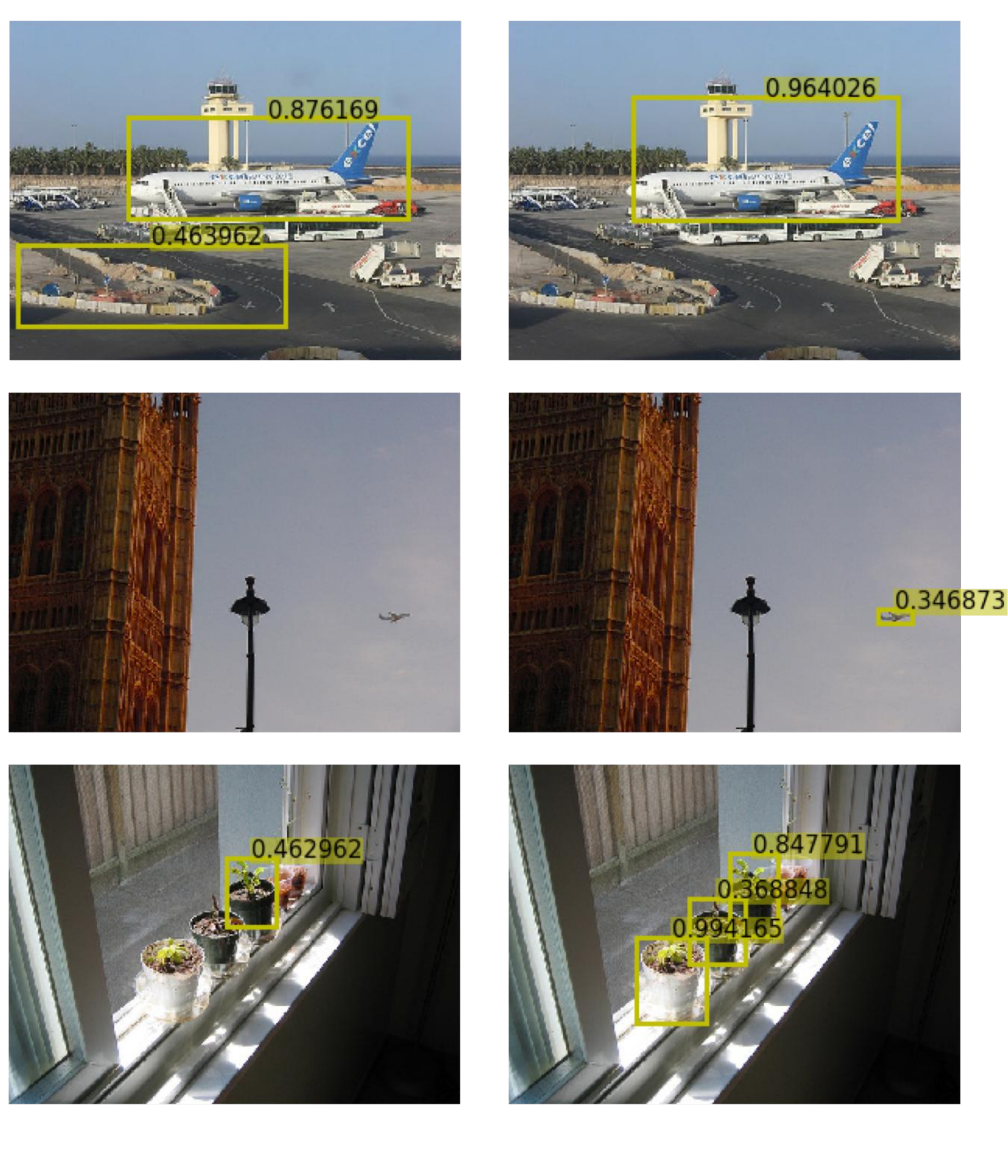}
 \caption{Examples of detection results. Left: Original SSD300. Right: DES300.
 See details in Section~\ref{sec:demo}.}
 \label{fig:demo}
\end{figure}

We show some detection examples in Figure~\ref{fig:demo}.
Left column is the
result of original SSD300, and the right column is the result of our DES300.
We show `aeroplane' in the first two rows, and `pottedplant' in the last row,
for all detection results with a score higher than 0.3.
From these examples, we can see that our method is good at detecting small
objects like small aeroplanes and pottedplants, and it can also prune out some
false positives which are incorrectly detected as aeroplane
in the first row.

\section{Conclusion}
In this paper, we propose a novel single shot object detector named
Detection with Enriched Semantics (DES).
To address the problem that low level detection feature map does not have
high level semantic information,
we introduce a segmentation branch, which utilize
the idea of weakly supervised semantic segmentation, to provide high semantic meaningful and
class-aware features to activate and calibrate feature map used in the object
detection.
We also utilize a global activation module to
provide global context information and pure channel-wise learning.
Our method is flexible and simple, and does not
require too much modifications to the original detection framework SSD.
Quantitative evaluation shows our method excels in both accuracy and speed.
Our method can also be applied to other two-stage or single shot object detectors,
with stronger backbone, and we remain this as future work.

\section*{Acknowledgements}
\label{Acknowledgements}
This research was supported by
ONR grant N00014-15-1-2356, the Center for Brains, Minds, and Machines (CBMM),
funded by NSF STC award CCF-1231216, and the National Natural Science Foundation of China No. 61672336.

{
\small
\bibliographystyle{ieee}
\bibliography{egbib}
}

\end{document}